\documentclass[10pt,twocolumn,letterpaper]{article}
\usepackage{iccv}

\usepackage{times}
\usepackage{epsfig}
\usepackage{graphicx}
\usepackage{amsmath}
\usepackage{amssymb}
\usepackage{amsfonts}       
\usepackage{booktabs}       
\usepackage{multirow}
\usepackage{makecell}
\usepackage{algorithm,algorithmicx,algpseudocode}
\usepackage{pifont}

\usepackage[pagebackref=true,breaklinks=true,colorlinks,bookmarks=false]{hyperref}

\iccvfinalcopy 
\def\iccvPaperID{2930} 

\ificcvfinal\fi

\usepackage{amsmath,amsfonts,bm}

\def\eqref#1{equation~\ref{#1}}

\def\1{\bm{1}}

\def\vh{{\bm{h}}}

\def\vn{{\bm{n}}}

\def\vr{{\bm{r}}}
\def\vs{{\bm{s}}}

\def\vx{{\bm{x}}}
\def\vy{{\bm{y}}}

\def\mX{{\bm{X}}}
\def\mY{{\bm{Y}}}

\DeclareMathAlphabet{\mathsfit}{\encodingdefault}{\sfdefault}{m}{sl}
\SetMathAlphabet{\mathsfit}{bold}{\encodingdefault}{\sfdefault}{bx}{n}

\def\gE{{\mathcal{E}}}

\def\gL{{\mathcal{L}}}

\def\gN{{\mathcal{N}}}

\def\gS{{\mathcal{S}}}

\def\sR{{\mathbb{R}}}

\newcommand{\E}{\mathbb{E}}

\begin{document}

\title{Score-Based Point Cloud Denoising}

\author{Shitong Luo, Wei Hu\thanks{Corresponding author: Wei Hu (forhuwei@pku.edu.cn). This work was supported by National Natural Science Foundation of China (61972009).}\\
Wangxuan Institute of Computer Technology\\
Peking University\\
{\tt\small \{luost, forhuwei\}@pku.edu.cn}
}

\maketitle
\begin{abstract}
Point clouds acquired from scanning devices are often perturbed by noise, which affects downstream tasks such as surface reconstruction and analysis. 
The distribution of a noisy point cloud can be viewed as the distribution of a set of noise-free samples $p(\vx)$ convolved with some noise model $n$, leading to $(p * n)(\vx)$ whose mode is the underlying clean surface.
To denoise a noisy point cloud, we propose to increase the log-likelihood of each point from $p * n$ via gradient ascent---iteratively updating each point's position.
Since $p * n$ is unknown at test-time, and we only need the score (\ie, the gradient of the log-probability function) to perform gradient ascent, we propose a neural network architecture to estimate the score of $p * n$ given only noisy point clouds as input.
We derive objective functions for training the network and develop a denoising algorithm leveraging on the estimated scores.
Experiments demonstrate that the proposed model outperforms state-of-the-art methods under a variety of noise models, and shows the potential to be applied in other tasks such as point cloud upsampling. The code is available at \url{https://github.com/luost26/score-denoise}.

\end{abstract}

\vspace{-0.05in}
\section{Introduction}
\label{sec:intro}
\vspace{-0.05in}

Point clouds consist of discrete 3D points irregularly sampled from continuous surfaces. It is an increasingly popular representation widely applied in autonomous driving, robotics and immersive tele-presence.
However, point clouds are often perturbed by noise due to the inherent limitations of acquisition equipments or matching ambiguities in the reconstruction from images.
Noise in point clouds significantly affects downstream tasks such as rendering, reconstruction and analysis since the underlying structures are deformed.
Hence, point cloud denoising is crucial to relevant 3D vision applications.
Nevertheless, point cloud denoising is challenging due to the irregular and unordered characteristics of point clouds.

\begin{figure}
\begin{center}
    \includegraphics[width=0.47\textwidth]{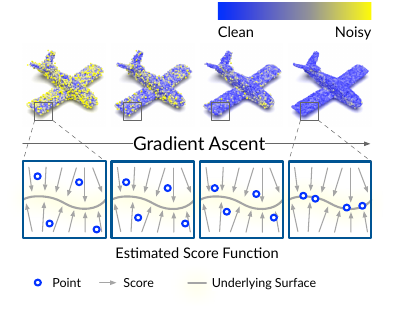}
\end{center}
\vspace{-0.25in}
   \caption{\textbf{An illustration of the proposed point cloud denoising method.} We first estimate the score of the noise-convolved distribution $\nabla_\vx \log[ (p * n)(\vx) ]$ from the noisy point cloud. Then, we perform gradient ascent using the estimated score to denoise the point cloud. }
\label{fig:teaser}
\vspace{-0.20in}
\end{figure}

Early point cloud denoising methods are optimization-based \cite{digne2017bilateral, huang2013bilat, cazals2005jetsfit, alexa2001MLS, avron2010sparsecoding, mattei2017MRPCA, sun2015lzero, zaman2017density}, which rely heavily on geometric priors and are sometimes challenging to strike a balance between the detail preservation and denoising effectiveness. 
Recently, deep-learning-based approaches have emerged and achieved promising denoising performance thanks to the advent of neural network architectures crafted for point clouds \cite{qi2017pointnet, qi2017pointnet2, wang2019dynamic}.
The majority of deep-learning-based denoising models predict the displacement of noisy points from the underlying surface and then apply the inverse displacement to the noisy point clouds \cite{duan2019NeuralProj, rakotosaona2020PCN, hermosilla2019TotalDenoising, pistilli2020learning}.  
This class of methods mainly suffer from two types of artifacts: shrinkage and outliers, which arise from over-estimation or under-estimation of the displacement.  
Instead, Luo \etal \cite{luo2020DMR} proposed to learn the underlying manifold of a noisy point cloud for reconstruction in a downsample-upsample architecture, which alleviates the issue of outliers by learning to filter out high-noise points in the downsampling stage. However, the downsampling stage inevitably causes detail loss especially at low noise levels.

In this paper, we propose a novel paradigm of point cloud denoising motivated by the distributional properties of noisy point clouds.
Point clouds consist of points sampled from the surface of 3D objects. Therefore, a noise-free point cloud can be modeled as a set of samples from some 3D distribution $p(\vx)$ supported by 2D manifolds.
If the point cloud is corrupted by noise, the distribution about the noisy point cloud can be modeled as the convolution between the original distribution $p$ and some noise model $n$ (\eg, Gaussian noise), expressed as $(p * n)(\vx)$.
Under some mild assumptions about the noise model $n$ (see Section~\ref{sec:analysis} for details), the mode of $p * n$ is the underlying clean surface, having higher probability than its ambient space.
According to this observation, denoising a noisy point cloud naturally amounts to moving noisy points towards the mode, which can be realized by performing gradient ascent on the log-probability function $\log[(p * n)(\vx)]$, as illustrated in Figure~\ref{fig:teaser}. 
As the points are expected to converge to the mode of distribution after sufficient iterations of gradient ascent, our method is more robust against artifacts such as shrinkage and outliers, while previous methods have no awareness of the mode.

However, there is a major challenge to address in order to implement this method---$p * n$ is unknown at test-time, which has to be estimated from the input noisy point cloud only.
To tackle this challenge, we propose a detail-preserving neural network architecture to estimate the {\it score} of the distribution underlying an input noisy point cloud $\nabla_\vx \log[(p * n)(\vx)]$, \ie, the gradient of the log-probability function.
We also formulate the objective function for training the score estimation network and develop a denoising algorithm.
Further, we provide an analysis of the model from the perspective of probability, revealing the principle behind the model formally.
Extensive experiments demonstrate that the proposed model outperforms state-of-the-art methods, and has the potential to be applied to other tasks such as point cloud upsampling.

To summarize, the contributions of this work include:
\vspace{-1.5mm}
\begin{itemize}
\setlength{\itemsep}{0pt}
\setlength{\parskip}{0pt}
    \item We propose a novel paradigm of point cloud denoising, exploiting the distribution model of noisy point clouds and leveraging the score of the distribution.
    
    \item To implement the method, we propose a neural network architecture for score-estimation from noisy point clouds, formulate objective functions for training the network, and develop the denoising algorithm.
    
    \item Extensive experiments demonstrate the capability of the proposed method under a variety of noise models.
    
\end{itemize}

\vspace{-0.05in}
\section{Related Work}
\label{sec:related}
\vspace{-0.05in}

\begin{figure}
\begin{center}
\includegraphics[width=0.5\textwidth]{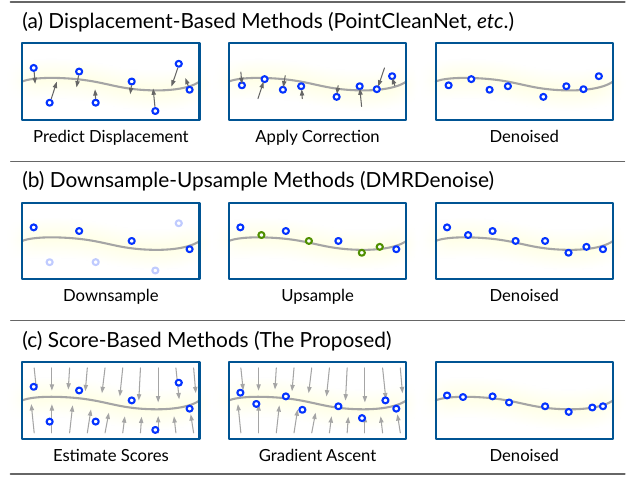}
\end{center}
   \caption{Illustration of different classes of deep-learning-based point cloud denoising methods.}
\label{fig:principle}
\end{figure}

\subsection{Optimization-based denoising}
\vspace{-0.05in}
Prior to the emergence of deep-learning-based denoising, the point cloud denoising problem is often formulated as an optimization problem constrained by geometric priors.
We classify them into four categories: 
(1) \textbf{Density-based} methods are most relevant to ours as they also involve modeling the distribution of points. \cite{zaman2017density} uses the kernel density estimation technique to approximate the density of noisy point clouds. Then, it removes outlying points in low-density regions. To finally obtain a clean point cloud, it relies on the bilateral filter \cite{fleishman2003bilateral} to reduce the noise of the outlier-free point cloud. Therefore, this method focuses on outlier removal.
(2) \textbf{Local-surface-fitting-based} methods approximate the point cloud with a smooth surface using simple-form function approximators and then project points onto the surface \cite{alexa2001MLS}. \cite{fleishman2003bilateral, cazals2005jetsfit, huang2013bilat, digne2017bilateral} proposed jet fitting and bilateral filtering that take into account both point coordinates and normals.
(3) \textbf{Sparsity-based} methods first reconstruct normals by solving an optimization problem with sparse regularization and then update the coordinates of points based on the reconstructed normals \cite{avron2010sparsecoding, sun2015lzero, xu2015sparsity}. The recently proposed MRPCA \cite{mattei2017MRPCA} is a sparsity-based denoiser which has achieved promising denoising performance.
(4) \textbf{Graph-based} methods represent point clouds on graphs and perform denoising using graph filters such as the graph Laplacian \cite{sch2015graphbased,zeng2019GLR, hu2020featuregraph, Hu2020gsp, hu20dynamic}. Recently, \cite{zeng2019GLR} proposed graph Laplacian regularization (GLR) of a low-dimensional manifold model for point cloud denoising, while \cite{hu2020featuregraph} proposed a paradigm of feature graph learning to infer the underlying graph structure of point clouds for denoising.
To summarize, optimization-based point cloud denoising methods rely heavily on geometric priors. Also, there is sometimes a trade-off between detail preservation and denoising effectiveness.

\vspace{-0.05in}
\subsection{Deep-learning-based denoising}
\vspace{-0.05in}
The advent of point-based neural networks \cite{qi2017pointnet, qi2017pointnet2, wang2019dynamic} has made deep point cloud denoising possible. 
The majority of existing deep learning based methods predict the displacement of each point in noisy point clouds using neural networks, and apply the inverse displacement to each point as illustrated in Figure~\ref{fig:principle}(a). 
PointCleanNet (PCNet) \cite{rakotosaona2020PCN} is the pioneer of this class of approaches, which employs a variant of PointNet as its backbone network. 
\cite{pistilli2020learning} proposed GPDNet, which uses graph convolutional networks to enhance the robustness of the neural denoiser.
\cite{hermosilla2019TotalDenoising} proposed an unsupervised point cloud denoising framework---Total Denoising (TotalDn).
In TotalDn, an unsupervised loss function is derived for training deep-learning-based denoisers, based on the assumption that points with denser surroundings are closer to the underlying surface.
The aforementioned displacement-prediction methods generally suffer from two types of artifacts: shrinkage and outliers, as a result of inaccurate estimation of noise displacement.
Instead, \cite{luo2020DMR} proposed to learn the underlying manifold (surface) of a noisy point cloud for reconstruction in a downsample-upsample architecture as illustrated in Figure~\ref{fig:principle}(b).
However, although the downsampling stage discards outliers in the input, it may also discard some informative details, leading to over-smoothing.

In this work, we propose a novel framework that distinguishes significantly from the aforementioned methods. Our method is motivated by the distribution model of noisy point clouds. It denoises point clouds via gradient ascent guided by the estimated gradient of the noisy point cloud's log-density as illustrated in Figure~\ref{fig:principle}(c).
Our method is shown to alleviate the artifacts of shrinkage and outliers, and achieve significantly better denoising performance.

\vspace{-0.05in}
\subsection{Score matching}
\vspace{-0.05in}
Score matching is a technique for training energy-based models---a family of non-normalized probability distributions \cite{lecun2006tutorial}.
It deals with matching the model-predicted gradients and the data log-density gradients by minimizing the squared distance between them \cite{hyvarinen2005estimation, song2019generative}.
Our proposed training objectives are similar to the score matching technique. The score matching technique in generative modeling aims at approximating unconditional distributions about data (\eg, images), while our model estimates the noise-convolved distribution of points.

Score matching has been applied to developing generative models for 3D shapes. 
\cite{cai2020shapegf} proposed an auto-encoder architecture ShapeGF that also has a score-estimation network which served as a decoder.
However, ShapeGF is different from our model in at least the following three aspects.
First, ShapeGF is designed for 3D point cloud generation and models the noise-free 3D distribution $p$, while our method models the noise-convolved distribution $p * n$ and aims at denoising the point cloud based on the score of $p * n$.
Second, since ShapeGF is a general auto-encoder for 3D shapes, it does not have the generalizability to out-of-distribution shapes. For instance, when trained on the ShapeNet dataset \cite{chang2015shapenet}, it can hardly generalize to shapes beyond the categories in ShapeNet. 
In contrast, our model is generalizable to arbitrary 3D shapes because our score function is defined on a local basis, which learns the building blocks of 3D shapes rather than the entire shapes themselves. This way narrows down the latent space of 3D geometry representation and makes it possible for the network to learn and reconstruct 3D details.
Third, to recover 3D shapes, ShapeGF requires a latent code of the shape obtained from the encoder, but their encoder is not meant to learn representations for denoising or other detail-demanding tasks.

\vspace{-0.05in}
\section{Method}
\label{sec:method}
\vspace{-0.05in}

\begin{figure}
\begin{center}
    \includegraphics[width=0.5\textwidth]{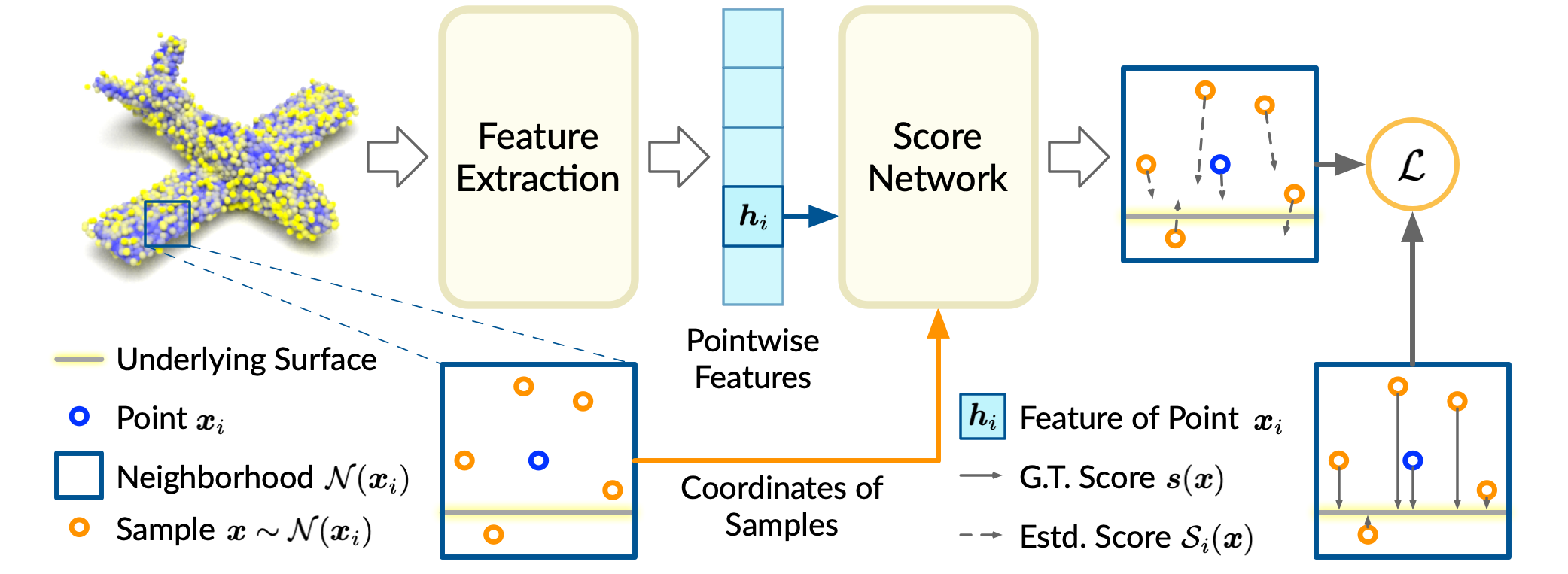}
\end{center}
\vspace{-0.1in}
  \caption{Illustration of the proposed network architecture.}
\label{fig:network}
\end{figure}

We first provide an overview of the proposed method. Then, we elaborate on the score estimation network, propose the training objective for the network, and develop a score-based denoising algorithm. 

\vspace{-0.05in}
\subsection{Overview}
\vspace{-0.05in}
Given a \emph{noisy} point cloud $\mX = \{ \vx_i \}_{i=1}^{N}$ consisting of $N$ points as input, we model the underlying \emph{noise-free} point cloud as a set of samples from a 3D distribution $p$ supported by 2D manifolds, and assume the noise follows a distribution $n$. Then the distribution of the noisy point cloud can be modelled as the convolution between $p$ and $n$, denoted as $p * n$.  

In order to denoise the noisy input $\mX$, we propose to estimate the score of the noise-convolved distribution $p * n$, \ie, $\nabla_\vx \log[(p * n)(\vx)]$---the gradient of the log-probability function, only from $\mX$. Then, we denoise $\mX$ using the estimated scores of $p * n$ via gradient ascent, thus moving noisy points towards the mode of the distribution that corresponds to the underlying clean surface.
The implementation of the proposed method mainly consists of the following three parts:
\vspace{-2mm}
\begin{enumerate}
\setlength{\itemsep}{0pt}
\setlength{\parskip}{0pt}
    \item \textbf{The score estimation network} that takes noisy point clouds as input and outputs point-wise scores $\nabla_\vx \log[(p * n)(\vx_i)] (i = 1, \ldots, N) $ (Section~\ref{sec:network}).
    \item \textbf{The objective function} for training the score estimation network (Section~\ref{sec:objective}).
    \item \textbf{The score-based denoising algorithm} that leverages on the estimated scores to denoise point clouds (Section~\ref{sec:denoise}).
\end{enumerate}

\vspace{-0.05in}
\subsection{The Proposed Score Estimation Network}
\label{sec:network}
\vspace{-0.05in}

Given a noisy point cloud $\mX = \{ \vx_i \}_{i=1}^{N}$ as input, the score estimation network predicts $\nabla_\vx \log[(p * n)(\vx_i)]$ for each point in $\mX$.
We estimate the score for each point $\vx_i$ on a \emph{local} basis, \ie, the network aims at estimating the score function in the neighborhood space around $\vx_i$, denoted as $\gS_i(\vr)$.
Localized score functions are fundamental to the model's {\it generalizability} because in this way the model focuses on the basic fragments of 3D shapes rather than the entire shapes themselves, narrowing down the latent space of 3D geometry representation.

The estimation of $\gS_i(\vr)$ is realized by a neural network which consists of a \emph{feature extraction unit} and a \emph{score estimation unit}. 
The feature extraction unit produces features that encode both the local and non-local geometry at each point. The extracted features are subsequently fed as parameters into the score estimation unit to construct score functions.

\paragraph{Feature Extraction Unit}
The feature extraction unit aims to learn point-wise features from the input noisy point cloud $\mX = \{ \vx_i\}_{i=1}^N$.
We adopt the feature extraction network widely used in previous denoising and upsampling models \cite{luo2020DMR, wang2019MPU, li2019PUGAN}.
Specifically, we construct a stack of densely connected dynamic graph convolutional layers \cite{wang2019dynamic}. 
The dynamic graph convolution is able to extract multi-scale as well as both local and non-local features for each point, while the dense connection produces features with richer contextual information \cite{huang2017densely, liu2019densepoint}.
These properties make the architecture suitable for the denoising task, as evidenced in previous works \cite{luo2020DMR, wang2019MPU}.
The learned feature for point $\vx_i$ is denoted as $\vh_i$. 

\paragraph{Score Estimation Unit}
The score estimation unit is parameterized by point $\vx_i$'s feature $\vh_i$.
It takes some 3D coordinate $\vx \in \sR^3$ nearby $\vx_i$ as input and outputs the score $\gS_i(\vx)$. Note that, here $\vx$ does not necessarily correspond to a point in the input point cloud $\mX$. It might be an intermediate coordinate during the gradient ascent denoising process.
Formally, the score estimation unit takes the form:
\begin{equation}
\label{eq:gnet}
    \gS_i(\vx) = \operatorname{Score}(\vx - \vx_i, \vh_i),
\end{equation}
where $\operatorname{Score}(\cdot)$ is a multi-layer perceptron (MLP). 
Note that we input $\vx - \vx_i$ (the coordinate of $\vx$ relative to $\vx_i$) to the network because the score function is localized around $\vx_i$.

The score estimation is trained by optimizing the proposed objective, which will be discussed next. 

\vspace{-0.05in}
\subsection{The Proposed Training Objective}
\label{sec:objective}
\vspace{-0.05in}

We denote the input noisy point cloud as $\mX = \{ \vx_i\}_{i=1}^N$ and the ground truth noise-free point cloud as $\mY = \{ \vy_i \}_{i=1}^N$.
Using the ground truth $\mY$, we define the score for some point $\vx \in \sR^3$ as follows:
\begin{equation}
\label{eq:gtscore}
    \vs(\vx) = \operatorname{NN}(\vx, \mY) - \vx, \ \vx \in \sR^3,
\end{equation}
where $\operatorname{NN}(\vx, \mY)$ returns the point nearest to $\vx$ in $\mY$.
Intuitively, $\vs(\vx)$ is a vector from $\vx$ to the underlying surface.

The training objective aligns the network-predicted score to the ground truth score defined above:
\begin{equation}
\label{eq:supervised}
    \mathcal{L}^{(i)} = \E_{\vx \sim \gN(\vx_i)} \left[ \left\| \vs(\vx) - \gS_{i}(\vx)  \right\|_2^2 \right],
\end{equation}
where $\gN(\vx_i)$ is a distribution concentrated in the neighborhood of $\vx_i$ in $\sR^3$ space.
Note that, this objective not only matches the predicted score on the position of $\vx_i$ but also matches the score on the neighboring areas of $\vx_i$ as illustrated in Figure~\ref{fig:network}.
This is important because a point moves around during gradient ascent, which relies on the score defined on the neighborhood of its initial position.
Such definition of training objective also distinguishes our method from previous displacement-based methods \cite{rakotosaona2020PCN, pistilli2020learning}, as the objectives of those methods only consider the position of each point while our proposed objective covers the neighborhood of each point.

The final training objective is simply an aggregation of the objective for each local score function:
\begin{equation}
    \mathcal{L} = \frac{1}{N}\sum_{i=1}^{N} \mathcal{L}^{(i)}.
\end{equation}

\vspace{-0.05in}
\subsection{The Score-Based Denoising Algorithm}
\label{sec:denoise}
\vspace{-0.05in}

Given a noisy point cloud $\mX = \{ \vx_i \}_{i=1}^N$ as input, we first need to construct the local score function $\gS_i$ for each point $\vx_i \in \mX$.
Specifically, we first feed the input point cloud $\mX$ to the feature extraction unit to obtain a set of point-wise features $\{ \vh_i \}_{i=1}^N$.
Next, by substituting $\vx_i$, $\vh_i$ and some 3D coordinate $\vx \in \sR^3$ into Eq.~\ref{eq:gnet}, we obtain $\gS_i(\vx)$ as the estimated score at $\vx$.

In principle, we can solely use $\gS_i$ to denoise $\vx_i$. However, to enhance the robustness and reduce the bias of estimation, we propose the ensemble score function:
\begin{equation}
\label{eq:ensemble}
    \gE_i(\vx) = \frac{1}{K} \sum_{\vx_j \in k\operatorname{NN}(\vx_i)} \gS_j(\vx), \ \vx \in \sR^3,
\end{equation}
where $k\operatorname{NN}(\vx_i)$ is $\vx_i$'s $k$-nearest neighborhood.

Finally, denoising a point cloud amounts to updating each point's position via gradient ascent:
\begin{equation}
\label{eq:denoise}
\begin{split}
    \vx_i^{(t)} & = \vx_i^{(t-1)} + \alpha_t \gE_i(\vx_i^{(t-1)}), \ t = 1,\ldots,T , \\
    \vx_i^{(0)} & = \vx_i, \ \vx_i \in \mX ,
\end{split}
\end{equation}
where $\alpha_t$ is the step size at the $t$-th step.
We suggest two criteria for choosing the step size sequence $\{ \alpha_t \}_{t=1}^T$: 
(1) The sequence should be decreasing towards 0 to ensure convergence. 
(2) $\alpha_1$ should be less than 1 and not be too close to 0, because according to Eq.~\ref{eq:gtscore}, the magnitude of the score is approximately the distance from each point to the underlying surface (approximately the length of $\vs(\vx)$ in Eq.~\ref{eq:gtscore}). Thus, performing gradient ascent for a sufficient number of steps with a proper step size less than 1 is enough and avoids over-denoising.

It is worth noting that, unlike some previous deep-learning-based denoisers such as PCNet \cite{rakotosaona2020PCN} and TotalDn \cite{hermosilla2019TotalDenoising} that suffer from shape shrinkage, we do not observe any shrinkage induced by our method. 
Thus, we have no need to post-process the denoised point clouds by inflating them slightly as in those works.  
This shows that our method is more robust to shrinkage compared to previous ones.

\vspace{-0.05in}
\section{Analysis}
\label{sec:analysis}
\vspace{-0.05in}

\begin{figure}
\begin{center}
    \includegraphics[width=0.45\textwidth]{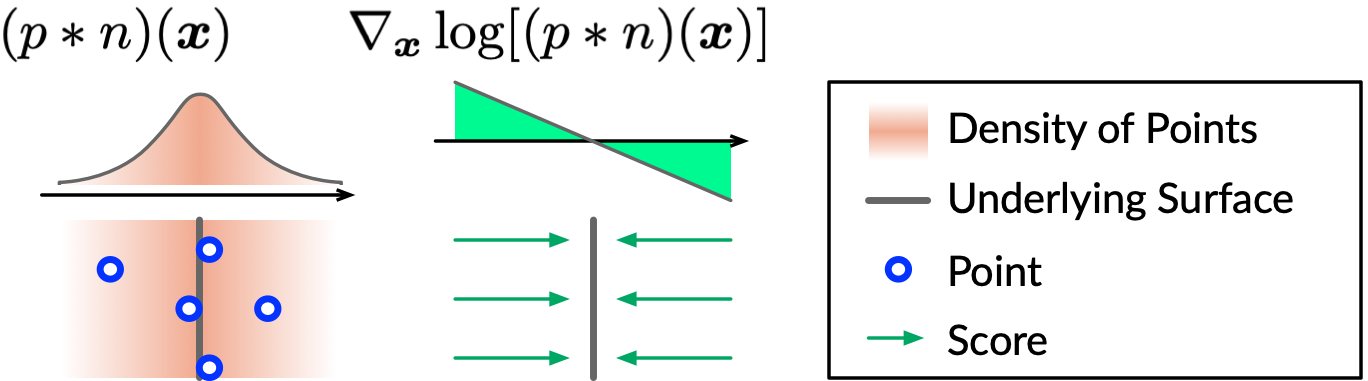}
\end{center}
\vspace{-0.05in}
   \caption{Illustration of the denoising theory.}
\label{fig:denoising_theory}
\end{figure}

\begin{table*}
\begin{center}
\resizebox{\textwidth}{!}{
\begin{tabular}{c|l| cccccc|cccccc}
\toprule
\multicolumn{2}{c|}{\# Points} & \multicolumn{6}{c|}{10K (Sparse)} & \multicolumn{6}{c}{50K (Dense)} \\

\multicolumn{2}{c|}{Noise} & 
\multicolumn{2}{c}{1\%} & \multicolumn{2}{c}{2\%} & \multicolumn{2}{c|}{3\%} &
\multicolumn{2}{c}{1\%} & \multicolumn{2}{c}{2\%} & \multicolumn{2}{c}{3\%} \\

Dataset & \makecell[c]{Model} &
 CD & P2M & CD & P2M & CD & P2M &
 CD & P2M & CD & P2M & CD & P2M \\
\midrule

\multirow{8}{*}{PU \cite{yu2018PUNet}} 
    & Bilateral \cite{fleishman2003bilateral} &
        3.646 & 1.342 & 5.007 & 2.018 & 6.998 & 3.557 &
        0.877 & 0.234 & 2.376 & 1.389 & 6.304 & 4.730
        \\
    & Jet \cite{cazals2005jetsfit}   &
        2.712 & 0.613 & 4.155 & 1.347 & 6.262 & 2.921 &
        0.851 & 0.207 & 2.432 & 1.403 & 5.788 & 4.267 
        \\
    & MRPCA \cite{mattei2017MRPCA} &
        2.972 & 0.922 & 3.728 & 1.117 & 5.009 & 1.963 &
        \bf 0.669 & \bf 0.099 & 2.008 & 1.033 & 5.775 & 4.081 
        \\
    & GLR \cite{zeng2019GLR}   &
        2.959 & 1.052 & 3.773 & 1.306 & 4.909 & 2.114 &
        0.696 & 0.161 & 1.587 & 0.830 & 3.839 & 2.707 
        \\
\cmidrule{2-14}
    & PCNet \cite{rakotosaona2020PCN}   &
        3.515 & 1.148 & 7.467 & 3.965 & 13.067 & 8.737 &
        1.049 & 0.346 & 1.447 & 0.608 & 2.289 & 1.285 
        \\
    & DMR \cite{luo2020DMR}  &
        4.482 & 1.722 & 4.982 & 2.115 & 5.892 & 2.846 &
        1.162 & 0.469 & 1.566 & 0.800 & 2.432 & 1.528 
        \\
\cmidrule{2-14}
    & \bf Ours &
        \bf 2.521 & \bf 0.463 & \bf 3.686 & \bf 1.074 & \bf 4.708 & \bf 1.942 &
        \bf 0.716 & \bf 0.150 & \bf 1.288 & \bf 0.566 & \bf 1.928 & \bf 1.041 
        \\
    
\midrule\midrule

\multirow{8}{*}{PC \cite{rakotosaona2020PCN}} 
    & Bilateral \cite{fleishman2003bilateral} & 
        4.320 & 1.351 & 6.171 & 1.646 & 8.295 & 2.392 &
        1.172 & 0.198 & 2.478 & 0.634 & 6.077 & 2.189 
        \\
    & Jet \cite{cazals2005jetsfit}   & 
        \bf 3.032 & \bf 0.830 & 5.298 & 1.372 & 7.650 & 2.227 &
        1.091 & 0.180 & 2.582 & 0.700 & 5.787 & 2.144 
        \\
    & MRPCA \cite{mattei2017MRPCA} & 
        3.323 & 0.931 & \bf 4.874 & 1.178 & \bf 6.502 & 1.676 &
        0.966 & 0.140 & 2.153 & 0.478 & 5.570 & 1.976
        \\
    & GLR \cite{zeng2019GLR}   & 
        3.399 & 0.956 & 5.274 & \bf 1.146 & 7.249 & \bf 1.674 &
        \bf 0.964 & \bf 0.134 & 2.015 & 0.417 & 4.488 & 1.306 
        \\
\cmidrule{2-14}
    & PCNet \cite{rakotosaona2020PCN}   & 
        3.847 & 1.221 & 8.752 & 3.043 & 14.525 & 5.873 &
        1.293 & 0.289 & 1.913 & 0.505 & 3.249 & 1.076 
        \\
    & DMR \cite{luo2020DMR}  & 
        6.602 & 2.152 & 7.145 & 2.237 & 8.087 & 2.487 &
        1.566 & 0.350 & 2.009 & 0.485 & 2.993 & 0.859 
        \\
\cmidrule{2-14}
    & \bf Ours & 
        \bf 3.369 & \bf 0.830 & \bf 5.132 & \bf 1.195 & \bf 6.776 & \bf 1.941 &
        \bf 1.066 & \bf 0.177 & \bf 1.659 & \bf 0.354 & \bf 2.494 & \bf 0.657 
        \\
    
\bottomrule

\end{tabular}
}
\end{center}
\caption{Comparison among competitive denoising algorithms. CD is multiplied by $10^4$ and P2M is multiplied by $10^4$.}
\label{table:quantitative}
\end{table*}

In this section, we elaborate on the distribution model for noisy point clouds.

\subsection{Points as Samples from a Distribution}
\vspace{-0.05in}

To begin with, we consider the distribution of a noise-free point cloud $\mY = \{\vy_i\}_{i=1}^N$ as sampled from a 3D distribution $p(\vy)$ supported by 2D manifolds. 
Since $p(\vy)$ is supported on 2D manifolds, it is discontinuous and has zero support in the ambient space, \ie, $p(\vy) \rightarrow \infty$ if $\vy$ exactly lies on the manifold, otherwise $p(\vy) = 0$.

Next, we consider the distribution of noisy point clouds.
A noisy point cloud can be denoted as $\mX = \{ \vx_i = \vy_i + \vn_i \}_{i=1}^{N}$, where $\vn_i$ is the noise component from a distribution $n$.
Here, we assume that the probability density function $n$ is continuous and has a unique mode at 0.
These assumptions are made for analysis. We will show by experiments that in some cases where the assumptions do not hold, the proposed method still achieves superior performance (see Section A in the supplementary material).
Under the continuity assumption of $n$, the density function of the distribution with respect to $\vx_i$ can be expressed as a convolution of $p$ and $n$:
\begin{equation}
\begin{split}
    q(\vx) & := (p * n)(\vx) \\
    & = \int_{\vs \in \sR^3} p(\vs) n(\vx - \vs) \mathrm{d} \vs .
\end{split}
\end{equation}
It can be shown by taking the derivative of both sides that the noise-free point cloud $\mY$ from the noise-free distribution $p$ exactly lies on the mode of $q$ if the mode of $n$ is 0. 
When the assumption of uni-modality holds, $q(\vx)$ reaches the maximum on the noise-free manifold.

\vspace{-0.05in}
\subsection{Connection to Denoising}
\vspace{-0.05in}
Suppose the density function $q(\vx)$ is known. 
Based on the above analysis, denoising a point cloud $\mX = \{ \vx_i \}_{i=1}^N$ amounts to maximizing $\sum_i \log q(\vx_i)$.
This can be naturally achieved by performing gradient ascent until the points converge to the mode of $q(\vx)$. 
The gradient ascent relies \emph{only} on the score function $\nabla_\vx \log q(\vx)$---the first-order derivative of the log-density function.
As shown in the previous subsection, $q(\vx)$ reaches the maximum on the underlying manifold under some mild assumptions. 
Hence, the vector field $\nabla_\vx \log q(\vx)$ consistently heads to the clean surface as demonstrated in Figure~\ref{fig:denoising_theory}.

However, the density $q$ is unknown during test-time. Instead of estimating $q$ from noisy observations, we only need the gradient of $\log q$ during the denoising, which is more tractable.
This motivates the proposed model---score-based denoising.

\vspace{-0.05in}
\subsection{Connection to the Training Objective}
\vspace{-0.05in}
The training objective defined in Eq.~\ref{eq:supervised} matches the predicted score to the ground truth score function. 
The magnitude of the estimated score may not exactly equal to that of the real score function. However, this is not an issue in the denoising task, since as long as the directions of estimated gradients are accurate, the points will converge to the underlying surface with sufficient number of steps at a suitable step size of gradient ascent.

\vspace{-0.05in}
\section{Experiments}
\label{sec:experiment}
\vspace{-0.05in}

\begin{table}
\begin{center}
\resizebox{0.4\textwidth}{!}{
\begin{tabular}{c|cc|cc|c}
\toprule
 & MRPCA & GLR & PCNet & DMR & Ours \\
\midrule
CD & 
 2.886 & 2.663 & 3.137 & 2.764 & \bf 2.616 \\
P2M & 
 1.933 & 1.920 & 2.142 & 1.910 & \bf 1.847 \\
\bottomrule
\end{tabular}

}
\end{center}
\caption{Comparison among different denoising methods tested on point clouds generated by simulated LiDAR scanning with realistic LiDAR noise, which is an unseen noise pattern to our denoiser since we train only on Gaussian noise. CD is multiplied by $10^4$ and P2M is multiplied by $10^4$.}
\label{table:blensor}
\end{table}

\subsection{Setup}
\label{sec:experiment:setup}
\vspace{-0.05in}

\paragraph{Datasets}
We collect 20 meshes for training from the training set of PU-Net \cite{yu2018PUNet} and use Poisson disk sampling to sample points from the meshes, at resolutions ranging from 10K to 50K points.
The point clouds are normalized into the unit sphere. Then, they are \emph{only perturbed by Gaussian noise} with standard deviation from 0.5\% to 2\% of the bounding sphere's radius.
Similar to previous works \cite{rakotosaona2020PCN, luo2020DMR}, point clouds are split into patches before being fed into the model. We set the patch size to be 1K.

For quantitative evaluation, we use two benchmarks: the test-set of PU-Net \cite{yu2018PUNet} (20 shapes) and the test-set of PointCleanNet (10 shapes) \cite{rakotosaona2020PCN}.
Similarly, we use Poisson disk sampling to sample point clouds from each shape, at resolution levels of 10K and 50K points.
The performance of our model is then evaluated using a variety of noise models, including isotropic Gaussian noise, simulated LiDAR noise, non-isotropic Gaussian noise, uni-directional noise, Laplace noise, uniform noise, and discrete noise.

Furthermore, we also use the \emph{Paris-rue-Madame} dataset \cite{serna2014paris} for visual evaluation, which is obtained from the real world using laser scanners.

\vspace{-0.1in}
\paragraph{Baselines}
We compare our method to state-of-the-art deep-learning-based denoisers and optimization-based denoisers.

Deep-learning-based denoisers include: PointCleanNet (PCNet) \cite{rakotosaona2020PCN}, and DMRDenoise (DMR) \cite{luo2020DMR}.
We exclude Total Denoising (TotalDn) \cite{hermosilla2019TotalDenoising} in our main experiments as TotalDn is based on unsupervised learning and it is unfair to compare supervised and unsupervised models explicitly. 
However, we will present an unsupervised adaptation of our model inspired by the training objective proposed by \cite{hermosilla2019TotalDenoising} in the supplementary material, and compare our unsupervised adaptation to TotalDn.

Optimization-based denoisers include bilateral filtering \cite{digne2017bilateral}, jet fitting \cite{cazals2005jetsfit}, MRPCA \cite{mattei2017MRPCA} and GLR \cite{zeng2019GLR}.

\vspace{-0.1in}
\paragraph{Metrics}
We employ two metrics commonly adopted in previous works to perform quantitative evaluation of our model: Chamfer distance (CD) \cite{fan2017pointsetgen} and point-to-mesh distance (P2M) \cite{ravi2020pytorch3d}.
Since the size of point clouds varies, we normalize the denoised results into the unit sphere before computing the metrics.

\vspace{-0.1in}
\paragraph{Hyper-parameters}
We use \emph{only one} set of hyper-parameters to train a unique model for all experimental settings, except for ablation studies.
Hyper-parameters including learning rates, denoising step sizes, network architectures, \etc, are provided in the supplementary material.
The code and data are available at \url{https://github.com/luost26/score-denoise}.

\begin{figure*}
\begin{center}
    \includegraphics[width=1.0\textwidth]{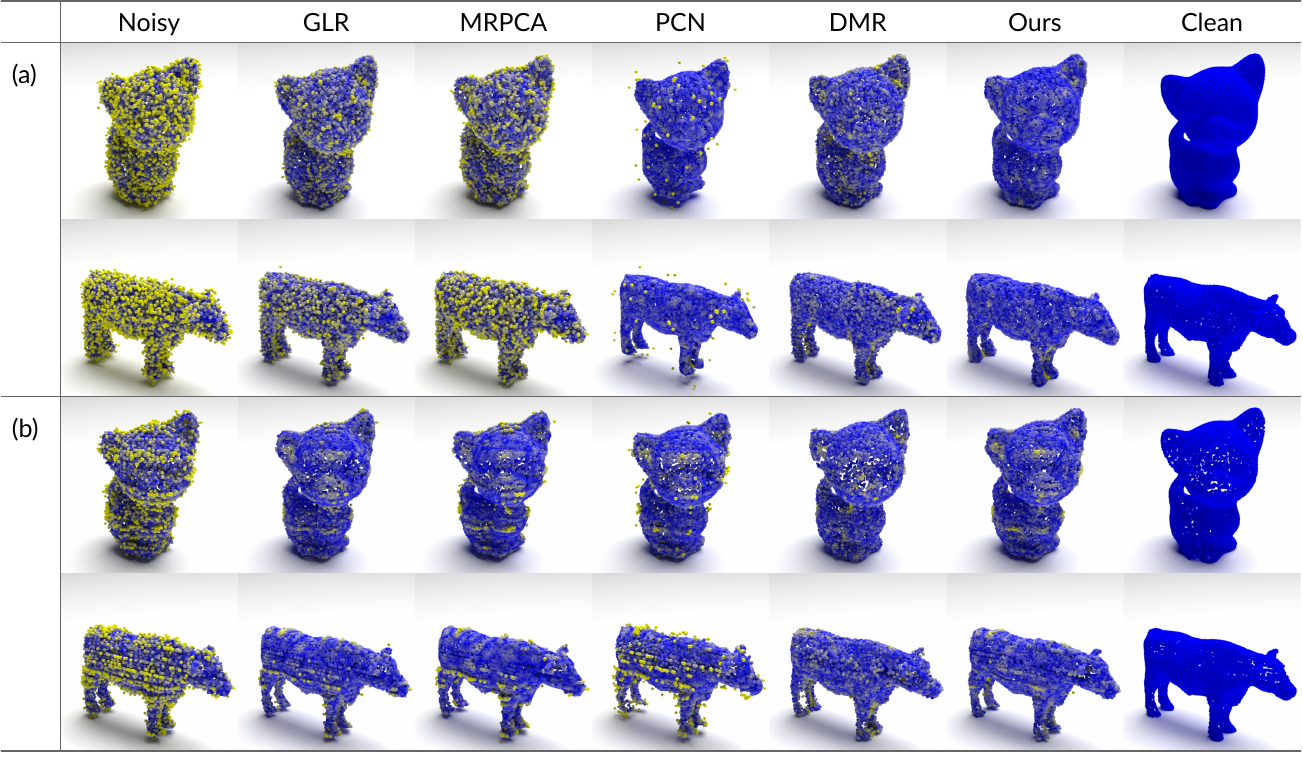}
\end{center}
\vspace{-0.05in}
\caption{Visual comparison of denoising methods under (a) Gaussian noise, (b) simulated LiDAR noise. Points colored yellower are farther away from the ground truth surface.}
\label{fig:visualization}
\end{figure*}

\vspace{-0.05in}
\subsection{Quantitative Results}
\label{sec:experiment:quant}
\vspace{-0.05in}

We first use isotropic Gaussian noise to test our models and baselines. The standard deviation of noise ranges from 1\% to 3\% of the shape's bounding sphere radius.
As presented in Table~\ref{table:quantitative}, our model significantly outperforms previous deep-learning-based methods in all settings and, surpasses optimization-based methods in the majority of cases.

Although the model is trained with only Gaussian noise, to test its generalizability, we use a different noise type---simulated LiDAR noise.
Specifically, we use a virtual Velodync HDL-64E2 scanner provided by the Blensor simulation package \cite{gschwandtner2011blensor} to acquire noisy point clouds. The scanning noise level is set to 1\% following \cite{luo2020DMR}.
The results in Table~\ref{table:blensor} indicate that although our denoiser is trained using Gaussian noise, it is effective in generalizing to unseen LiDAR noise and outperforms previous methods.

Other noise models, including non-isotropic Gaussian noise, uni-directional noise, Laplace noise, uniform noise, and discrete noise are also used to evaluate our method and baselines.
In most of these experimental settings, our model outperforms competing baselines.
The detailed results are included in the supplementary material.

\begin{figure*}
\begin{center}
\includegraphics[width=\textwidth]{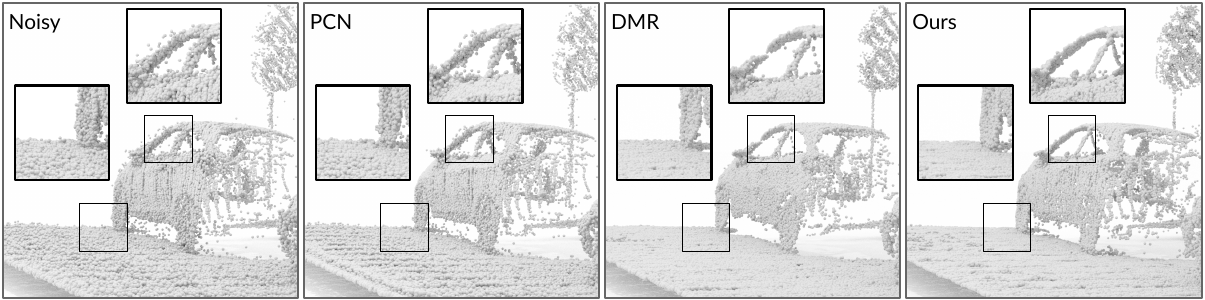}
\end{center}
\vspace{-0.1in}
\caption{Visual results of our denoiser on the real-world dataset \textit{Paris-rue-Madame} \cite{serna2014paris}.}
\label{fig:paris}
\end{figure*}

\vspace{-0.05in}
\subsection{Qualitative Results}
\vspace{-0.05in}
Figure~\ref{fig:visualization} shows the denoising results from the proposed method and competitive baselines under Gaussian noise and simulated LiDAR noise.
The color of each point indicates its reconstruction error measured by point-to-mesh distance introduced in Section~\ref{sec:experiment:setup}.
Points closer to the underlying surface are colored darker, otherwise colored brighter.
As can be observed in the figure, our results are much cleaner and more visually appealing than those of other methods.
Notably, our method preserves details better than other methods and is more robust to outliers compared to other deep-learning-based methods such as PCNet and DMRDenoise.

Further, we conduct qualitative studies on the real-world dataset \emph{Paris-rue-Madame} \cite{serna2014paris}. 
Note that, since the noise-free point cloud is unknown for real-world datasets, the error of each point cannot be computed and visualized.
As demonstrated in Figure~\ref{fig:paris}, our denoising result is cleaner and smoother than that of PCNet, with details preserved better than DMRDenoise.

\begin{figure}
\begin{center}
    \includegraphics[width=0.5\textwidth]{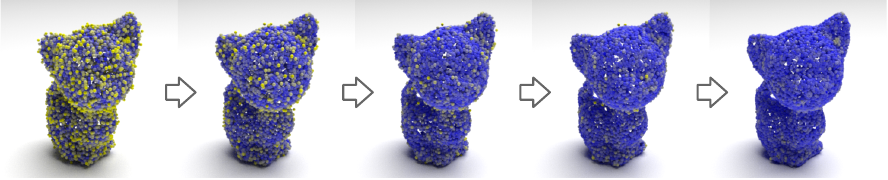}
\end{center}
\vspace{-0.1in}
   \caption{A gradient ascent trajectory of denoising.}
\label{fig:traj}
\end{figure}

In addition, we present a denoising trajectory in Figure~\ref{fig:traj}, which reveals the gradient ascent process of our method---noise reduces as points gradually converge to the mode of $p * n$.

More visual results regarding synthetic noise and real-world noise are provided in the supplementary material. 

In summary, the demonstrated qualitative results are consistent with the quantitative results in Section~\ref{sec:experiment:quant}, which again validates the effectiveness of the proposed method.

\vspace{-0.05in}
\subsection{Ablation Studies}
\vspace{-0.05in}
We perform ablation studies to assess the contribution of the proposed method's main designs:

\textbf{(1) Score-based denoising algorithm} We replace the gradient ascent rule (Eq.~\ref{eq:denoise}) by directly adding the predicted score to the input coordinates, which is similar to end-to-end displacement-based methods:
\begin{equation}
    \vy_i = \vx_i + \gE_i(\vx_i).
\end{equation}
We also apply this update rule iteratively following previous displacement-based methods \cite{rakotosaona2020PCN, hermosilla2019TotalDenoising}. The number of iterations is fine-tuned to produce the best performance.

\textbf{(2) Neighborhood-covering training objective} We replace the objective in Eq.~\ref{eq:supervised} with:
\begin{equation}
    \gL^{(i)} = \| \vs(\vx_i) - \gS_i(\vx_i) \|_2^2,
\end{equation}
which is similar to the L2 objective \cite{rakotosaona2020PCN} or the Chamfer distance \cite{luo2020DMR, pistilli2020learning} employed in previous deep-learning-based models \cite{rakotosaona2020PCN}, considering only the position of $\vx_i$, while ours covers the neighborhood of $\vx_i$.

\textbf{(3) Ensemble score function} We replace the ensemble score function in Eq.~\ref{eq:ensemble} with the single score function $\gS_i(\vx)$.
 
As shown in Table~\ref{table:ablation}, all the components contribute positively to the denoising performance. More results and analysis of the ablation studies can be found in the supplementary material.

\begin{table}
\begin{center}
\resizebox{0.5\textwidth}{!}{
    \begin{tabular}{l|cccccc}
\toprule
Dataset: PU & \multicolumn{2}{c}{10K, 1\%} & \multicolumn{2}{c}{10K, 2\%} & \multicolumn{2}{c}{10K, 3\%} \\
Ablation & CD & P2M & CD & P2M & CD & P2M \\
\midrule
\bf (1)         & 3.237 & 0.994 & 5.241 & 2.258 & 7.471 & 4.049 \\
\bf (1) + iter. & 3.237* & 0.994* & 5.241* & 2.258* & 6.073 & 2.953 \\
\bf (2) & 4.726 & 2.188 & 5.740 & 2.748 & 5.976 & 3.036 \\
\bf (3) & 2.522 & 0.471 & 4.021 & 1.280 & 6.872 & 3.497 \\
\midrule
\bf Full & \bf 2.521 & \bf 0.463 & \bf 3.686 & \bf 1.074 & \bf 4.708 & \bf 1.942 \\
\bottomrule
\end{tabular}

}
\end{center}
\caption{Ablation studies. CD is multiplied by $10^4$ and P2M is multiplied by $10^4$. (*) The best performance is achieved after running for only 1 iteration.}
\label{table:ablation}
\end{table}

\vspace{-0.05in}
\subsection{Beyond Denoising: Upsampling via Denoising}
\vspace{-0.05in}

Going beyond denoising, we show that the proposed method is applicable to point cloud upsampling.
In particular, given a sparse point cloud with $N$ points as input, we perturb it with Gaussian noise independently for $r$ times, leading to a noisy dense point cloud consisting of $rN$ points. Subsequently, we feed the noisy dense point cloud to our denoiser to acquire the final upsampled point cloud.

\begin{figure}
\begin{center}
    \includegraphics[width=0.5\textwidth]{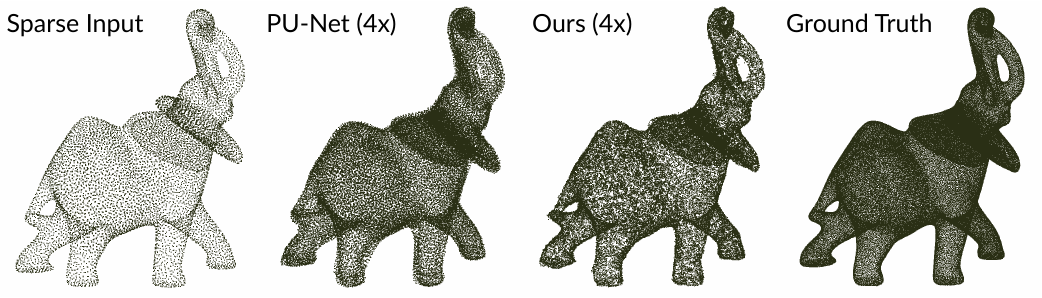}
\end{center}
\vspace{-0.05in}
   \caption{Visual comparison between the specialized upsampling method PU-Net and our denoising-based upsampling method.}
\label{fig:upsampling}
\end{figure}

We compare the denoising-based upsampling method with the classical upsampling network PU-Net \cite{yu2018PUNet} using the test-set of PU-Net. The quantitative results are presented in Table~\ref{table:upsampling_compare} and the qualitative comparison is shown in Figure~\ref{fig:upsampling}.
We see that the denoising-based upsampling method fairly outperforms PU-Net which is specialized in upsampling.
This implies that the proposed score-based method for point clouds has the potential in tasks beyond denoising, which will be further explored as our future works.

\begin{table}
\begin{center}
\small{
    \begin{tabular}{l|cc|cc}
\toprule
\#Points & \multicolumn{2}{c|}{5K} & \multicolumn{2}{c}{10K} \\
 & PU-Net \cite{yu2018PUNet} & \bf Ours & PU-Net \cite{yu2018PUNet} & \bf Ours \\
\midrule
CD & 3.445 & \bf 1.696 & 2.862 & \bf 1.454 \\
P2M & 1.669 & \bf 0.295 & 1.166 & \bf 0.181 \\
\bottomrule
\end{tabular}
}
\end{center}
\caption{Comparison between PU-Net and our denoising-based point cloud upsampling for the upsampling rate 4x. CD is multiplied by $10^4$ and P2M is multiplied by $10^4$.}
\label{table:upsampling_compare}
\end{table}
\vspace{-0.05in}
\section{Conclusions}
\vspace{-0.05in}

In this paper, we propose a novel paradigm of point cloud denoising, modeling noisy point clouds as samples from a noise-convolved distribution.
We design a neural network architecture to estimate the score of the distribution and leverage on the score to denoise point clouds via gradient ascent.
Experimental results validate the superiority of our model and further show the potential to be applied to other tasks such as point cloud upsampling.


{\small
\bibliographystyle{ieee_fullname.bst}
\bibliography{references}

\begin{thebibliography}{10}\itemsep=-1pt

\bibitem{alexa2001MLS}
Marc Alexa, Johannes Behr, Daniel Cohen-Or, Shachar Fleishman, David Levin, and
  Claudio~T Silva.
\newblock Point set surfaces.
\newblock In {\em Proceedings Visualization, 2001. VIS'01.}, pages 21--29.
  IEEE, 2001.

\bibitem{avron2010sparsecoding}
Haim Avron, Andrei Sharf, Chen Greif, and Daniel Cohen-Or.
\newblock $\ell$1-sparse reconstruction of sharp point set surfaces.
\newblock {\em ACM Transactions on Graphics (TOG)}, 29(5):1--12, 2010.

\bibitem{cai2020shapegf}
Ruojin Cai, Guandao Yang, Hadar Averbuch-Elor, Zekun Hao, Serge Belongie, Noah
  Snavely, and Bharath Hariharan.
\newblock Learning gradient fields for shape generation.
\newblock In {\em Proceedings of the European Conference on Computer Vision
  (ECCV)}, 2020.

\bibitem{cazals2005jetsfit}
Fr{\'e}d{\'e}ric Cazals and Marc Pouget.
\newblock Estimating differential quantities using polynomial fitting of
  osculating jets.
\newblock {\em Computer Aided Geometric Design}, 22(2):121--146, 2005.

\bibitem{chang2015shapenet}
Angel~X Chang, Thomas Funkhouser, Leonidas Guibas, Pat Hanrahan, Qixing Huang,
  Zimo Li, Silvio Savarese, Manolis Savva, Shuran Song, Hao Su, et~al.
\newblock Shapenet: An information-rich 3d model repository.
\newblock {\em arXiv preprint arXiv:1512.03012}, 2015.

\bibitem{digne2017bilateral}
Julie Digne and Carlo De~Franchis.
\newblock The bilateral filter for point clouds.
\newblock {\em Image Processing On Line}, 7:278--287, 2017.

\bibitem{duan2019NeuralProj}
Chaojing Duan, Siheng Chen, and Jelena Kovacevic.
\newblock 3d point cloud denoising via deep neural network based local surface
  estimation.
\newblock In {\em ICASSP 2019-2019 IEEE International Conference on Acoustics,
  Speech and Signal Processing (ICASSP)}, pages 8553--8557. IEEE, 2019.

\bibitem{fan2017pointsetgen}
Haoqiang Fan, Hao Su, and Leonidas~J Guibas.
\newblock A point set generation network for 3d object reconstruction from a
  single image.
\newblock In {\em Proceedings of the IEEE conference on computer vision and
  pattern recognition}, pages 605--613, 2017.

\bibitem{fleishman2003bilateral}
Shachar Fleishman, Iddo Drori, and Daniel Cohen-Or.
\newblock Bilateral mesh denoising.
\newblock In {\em ACM SIGGRAPH 2003 Papers}, pages 950--953. 2003.

\bibitem{gschwandtner2011blensor}
Michael Gschwandtner, Roland Kwitt, Andreas Uhl, and Wolfgang Pree.
\newblock Blensor: Blender sensor simulation toolbox.
\newblock In {\em International Symposium on Visual Computing}, pages 199--208.
  Springer, 2011.

\bibitem{hermosilla2019TotalDenoising}
Pedro Hermosilla, Tobias Ritschel, and Timo Ropinski.
\newblock Total denoising: Unsupervised learning of 3d point cloud cleaning.
\newblock In {\em Proceedings of the IEEE International Conference on Computer
  Vision}, pages 52--60, 2019.

\bibitem{hu2020featuregraph}
Wei Hu, Xiang Gao, Gene Cheung, and Zongming Guo.
\newblock Feature graph learning for 3{D} point cloud denoising.
\newblock {\em IEEE Transactions on Signal Processing}, 68:2841--2856, 2020.

\bibitem{hu20dynamic}
Wei Hu, Qianjiang Hu, Zehua Wang, and Xiang Gao.
\newblock Dynamic point cloud denoising via manifold-to-manifold distance.
\newblock {\em arXiv preprint arXiv:2003.08355}, 2020.

\bibitem{Hu2020gsp}
Wei Hu, Jiahao Pang, Xianming Liu, Dong Tian, Chia-Wen Lin, and Anthony Vetro.
\newblock Graph {S}ignal {P}rocessing for geometric data and beyond: Theory and
  applications.
\newblock {\em arXiv preprint arXiv:2008.01918}, 2020.

\bibitem{huang2017densely}
Gao Huang, Zhuang Liu, Laurens Van Der~Maaten, and Kilian~Q Weinberger.
\newblock Densely connected convolutional networks.
\newblock In {\em Proceedings of the IEEE conference on computer vision and
  pattern recognition}, pages 4700--4708, 2017.

\bibitem{huang2013bilat}
Hui Huang, Shihao Wu, Minglun Gong, Daniel Cohen-Or, Uri Ascher, and Hao Zhang.
\newblock Edge-aware point set resampling.
\newblock {\em ACM transactions on graphics (TOG)}, 32(1):1--12, 2013.

\bibitem{hyvarinen2005estimation}
Aapo Hyv{\"a}rinen.
\newblock Estimation of non-normalized statistical models by score matching.
\newblock {\em Journal of Machine Learning Research}, 6(Apr):695--709, 2005.

\bibitem{lecun2006tutorial}
Yann LeCun, Sumit Chopra, Raia Hadsell, M Ranzato, and F Huang.
\newblock A tutorial on energy-based learning.
\newblock {\em Predicting structured data}, 1(0), 2006.

\bibitem{li2019PUGAN}
Ruihui Li, Xianzhi Li, Chi-Wing Fu, Daniel Cohen-Or, and Pheng-Ann Heng.
\newblock Pu-gan: a point cloud upsampling adversarial network.
\newblock In {\em Proceedings of the IEEE International Conference on Computer
  Vision}, pages 7203--7212, 2019.

\bibitem{liu2019densepoint}
Yongcheng Liu, Bin Fan, Gaofeng Meng, Jiwen Lu, Shiming Xiang, and Chunhong
  Pan.
\newblock Densepoint: Learning densely contextual representation for efficient
  point cloud processing.
\newblock In {\em Proceedings of the IEEE International Conference on Computer
  Vision}, pages 5239--5248, 2019.

\bibitem{luo2020DMR}
Shitong Luo and Wei Hu.
\newblock Differentiable manifold reconstruction for point cloud denoising.
\newblock In {\em Proceedings of the 28th ACM International Conference on
  Multimedia}, pages 1330--1338, 2020.

\bibitem{mattei2017MRPCA}
Enrico Mattei and Alexey Castrodad.
\newblock Point cloud denoising via moving rpca.
\newblock In {\em Computer Graphics Forum}, volume~36, pages 123--137. Wiley
  Online Library, 2017.

\bibitem{pistilli2020learning}
Francesca Pistilli, Giulia Fracastoro, Diego Valsesia, and Enrico Magli.
\newblock Learning graph-convolutional representations for point cloud
  denoising.
\newblock {\em arXiv preprint arXiv:2007.02578}, 2020.

\bibitem{qi2017pointnet}
Charles~R Qi, Hao Su, Kaichun Mo, and Leonidas~J Guibas.
\newblock Pointnet: Deep learning on point sets for 3d classification and
  segmentation.
\newblock In {\em Proceedings of the IEEE conference on computer vision and
  pattern recognition}, pages 652--660, 2017.

\bibitem{qi2017pointnet2}
Charles~Ruizhongtai Qi, Li Yi, Hao Su, and Leonidas~J Guibas.
\newblock Pointnet++: Deep hierarchical feature learning on point sets in a
  metric space.
\newblock In {\em Advances in neural information processing systems}, pages
  5099--5108, 2017.

\bibitem{rakotosaona2020PCN}
Marie-Julie Rakotosaona, Vittorio La~Barbera, Paul Guerrero, Niloy~J Mitra, and
  Maks Ovsjanikov.
\newblock Pointcleannet: Learning to denoise and remove outliers from dense
  point clouds.
\newblock In {\em Computer Graphics Forum}, volume~39, pages 185--203. Wiley
  Online Library, 2020.

\bibitem{ravi2020pytorch3d}
Nikhila Ravi, Jeremy Reizenstein, David Novotny, Taylor Gordon, Wan-Yen Lo,
  Justin Johnson, and Georgia Gkioxari.
\newblock Accelerating 3d deep learning with pytorch3d.
\newblock {\em arXiv:2007.08501}, 2020.

\bibitem{sch2015graphbased}
Yann Schoenenberger, Johan Paratte, and Pierre Vandergheynst.
\newblock Graph-based denoising for time-varying point clouds.
\newblock In {\em 2015 3DTV-Conference: The True Vision-Capture, Transmission
  and Display of 3D Video (3DTV-CON)}, pages 1--4. IEEE, 2015.

\bibitem{serna2014paris}
Andr{\'e}s Serna, Beatriz Marcotegui, Fran{\c{c}}ois Goulette, and
  Jean-Emmanuel Deschaud.
\newblock Paris-rue-madame database: a 3d mobile laser scanner dataset for
  benchmarking urban detection, segmentation and classification methods.
\newblock 2014.

\bibitem{song2019generative}
Yang Song and Stefano Ermon.
\newblock Generative modeling by estimating gradients of the data distribution.
\newblock In {\em Advances in Neural Information Processing Systems}, pages
  11918--11930, 2019.

\bibitem{sun2015lzero}
Yujing Sun, Scott Schaefer, and Wenping Wang.
\newblock Denoising point sets via l0 minimization.
\newblock {\em Computer Aided Geometric Design}, 35:2--15, 2015.

\bibitem{wang2019dynamic}
Yue Wang, Yongbin Sun, Ziwei Liu, Sanjay~E Sarma, Michael~M Bronstein, and
  Justin~M Solomon.
\newblock Dynamic graph cnn for learning on point clouds.
\newblock {\em ACM Transactions on Graphics (TOG)}, 38(5):1--12, 2019.

\bibitem{xu2015sparsity}
Linlin Xu, Ruimin Wang, Juyong Zhang, Zhouwang Yang, Jiansong Deng, Falai Chen,
  and Ligang Liu.
\newblock Survey on sparsity in geometric modeling and processing.
\newblock {\em Graphical Models}, 82:160--180, 2015.

\bibitem{wang2019MPU}
Wang Yifan, Shihao Wu, Hui Huang, Daniel Cohen-Or, and Olga Sorkine-Hornung.
\newblock Patch-based progressive 3d point set upsampling.
\newblock In {\em Proceedings of the IEEE Conference on Computer Vision and
  Pattern Recognition}, pages 5958--5967, 2019.

\bibitem{yu2018PUNet}
Lequan Yu, Xianzhi Li, Chi-Wing Fu, Daniel Cohen-Or, and Pheng-Ann Heng.
\newblock Pu-net: Point cloud upsampling network.
\newblock In {\em Proceedings of the IEEE Conference on Computer Vision and
  Pattern Recognition}, pages 2790--2799, 2018.

\bibitem{zaman2017density}
Faisal Zaman, Ya~Ping Wong, and Boon~Yian Ng.
\newblock Density-based denoising of point cloud.
\newblock In {\em 9th International Conference on Robotic, Vision, Signal
  Processing and Power Applications}, pages 287--295. Springer, 2017.

\bibitem{zeng2019GLR}
Jin Zeng, Gene Cheung, Michael Ng, Jiahao Pang, and Yang Cheng.
\newblock 3{D} point cloud denoising using graph {Laplacian} regularization of
  a low dimensional manifold model.
\newblock {\em IEEE Transactions on Image Processing}, 29:3474--3489, December
  2019.

\end{thebibliography}
}

\end{document}